\documentclass[10pt,twocolumn,letterpaper]{article}
\usepackage[pagenumbers]{cvpr}
\usepackage{multirow}
\usepackage{arydshln}

\definecolor{cvprblue}{rgb}{0.21,0.49,0.74}
\usepackage[pagebackref,breaklinks,colorlinks,allcolors=cvprblue]{hyperref}

\def\confName{CVPR}
\def\confYear{2027}

\title{Beyond a Single Crop: In-Context Adaptation for Player Re-Identification}
\author{
Ge Luo$^{1}$\thanks{Equal contribution.}\quad
Changhuo Yang$^{1}$\footnotemark[1]\quad
Weisi Fan$^{2}$\\
Rui Yang$^{3}$\quad
Shaokai Wang$^{1}$\thanks{Corresponding author: \texttt{skwang@csu.edu.cn}.}\\
$^{1}$Central South University\\
$^{2}$Iowa State University\quad $^{3}$Shandong University\\
{\tt\small skwang@csu.edu.cn}
}

\hypersetup{
pdftitle={Beyond a Single Crop: In-Context Adaptation for Player Re-Identification},
pdfauthor={Ge Luo; Changhuo Yang; Weisi Fan; Rui Yang; Shaokai Wang}
}
\begin{document}
\maketitle

\begin{abstract}
Person re-identification is typically addressed by learning identity-discriminative embeddings from individual person crops. However, player re-identification in broadcast sports videos presents a unique challenge: players from the same team often share similar appearances. Conventional methods encode each player crop independently, overlooking the surrounding players and repeated observations that could help resolve these ambiguities.
Here, we propose ContextReID, a framework for visual in-context adaptation that represents a player by jointly considering other players in the same frame and observations from different frames of the broadcast sequence. We implement this framework with a vision-language backbone trained with contrastive objectives for player retrieval and identity matching across the supplied context. Without requiring parameter updates or context identity labels at inference, the trained model adapts the target embedding to the supplied visual context.
Experiments on SportsMOT, DanceTrack, and SoccerNet-GSR demonstrate superior retrieval performance compared with single-crop baselines. Cross-sport transfer experiments within SportsMOT further show that in-context adaptation can improve retrieval on sports excluded from training. These results highlight the empirical value of our method in utilizing spatial and temporal context without requiring identity labels or cross-frame identity associations at inference.
\end{abstract}

\section{Introduction}

\begin{figure}[t]
    \centering
    \includegraphics[width=\columnwidth]{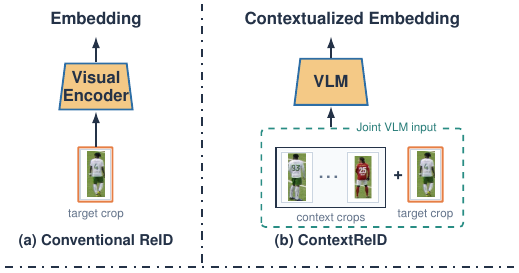}
    \caption{Visual in-context adaptation for player ReID. Conventional ReID encodes each player crop independently. ContextReID adapts the target embedding to supplied spatial and temporal player context, without parameter updates or context identity labels at inference. The output remains a retrieval vector for similarity-based matching.}
    \label{fig:framework-overview}
\end{figure}

Person re-identification (ReID) aims to retrieve observations of the same person across images or cameras \cite{ye2022survey}. Most approaches learn an identity-discriminative embedding from each person crop independently and compare the resulting vectors by similarity. This formulation supports efficient retrieval and has driven progress in metric learning, CNN architectures, and transformer-based representations \cite{hermans2017triplet,luo2020strong,he2021transreid}. However, independently encoding each crop leaves surrounding visual evidence unused. For player re-identification, that evidence can help determine which appearance cues distinguish a target from other players in the sequence.

Broadcast sports present a particular combination of ambiguity and context. Teammates often wear nearly identical uniforms, while distinguishing details such as jersey numbers, accessories, and body shape may be small, occluded, or blurred. Yet players are rarely observed in isolation: other players appear alongside them, and additional frames provide observations under different poses and viewpoints. A cue that is weak in isolation may become useful when interpreted relative to these alternatives. This motivates a question: can a trained retrieval model adapt a player's representation to the visual context of a sequence, without parameter updates or identity labels for that context?

We address this question through visual in-context adaptation. Inspired by in-context learning \cite{brown2020language}, we supply contextual evidence directly to the model and allow it to influence the target representation during the forward pass. Here, adaptation refers to a change in the embedding conditioned on the supplied visual input; the model parameters remain fixed at inference. The context contains unassociated player detections from the target frame and a small set of other frames. It requires neither identity labels nor cross-frame associations, and the target identity need not appear in the additional frames. This differs from aggregating observations from a known target tracklet or predicting identities from ID-labeled historical trajectories, as in MOTIP \cite{gao2025motip}.

We propose ContextReID, which implements this formulation with a vision-language backbone. The model jointly processes player crops organized by frame and produces a normalized retrieval embedding for the target crop. Training combines a contextual identity-matching objective with a retrieval objective, encouraging the model to use relationships within the supplied context while producing vectors suitable for cosine-similarity matching. We instantiate the framework with Qwen2-VL-2B-Instruct and Qwen3-VL-Embedding-2B \cite{wang2024qwen2vl,li2026qwen3vlembedding}. Figure~\ref{fig:framework-overview} contrasts independent crop encoding with this context-conditioned representation.

Experiments on SportsMOT, DanceTrack, and SoccerNet-GSR demonstrate that visual in-context adaptation improves player retrieval across two vision-language backbones. Cross-sport transfer experiments further show that contextual evidence can improve retrieval on sports excluded from task-specific training. Additional analyses examine when contextual information is useful and distinguish its contribution from the capabilities of the underlying pretrained models.

Our contributions are threefold:
\begin{itemize}
    \item We introduce ContextReID, a framework that adapts player embeddings to unlabeled spatial and temporal crop context at inference, without parameter updates, context identity labels, or cross-frame identity associations.
    \item We train two vision-language backbones with contextual identity-matching and retrieval objectives to produce context-adapted player embeddings.
    \item We demonstrate retrieval improvements on three benchmarks and under cross-sport transfer, with controlled analyses supporting the role of contextual evidence.
\end{itemize}

\section{Related Work}

\begin{figure*}[t]
    \centering
    \includegraphics[width=\textwidth]{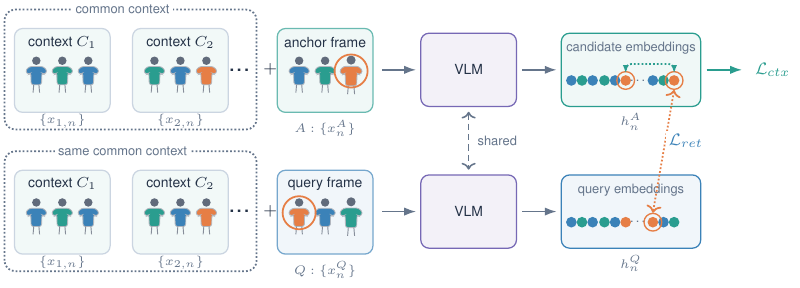}
    \caption{Detailed method. The anchor and query passes append different target frames to the same identity-free context prefix. The shared backbone produces contextual candidate and query embeddings, optimized jointly by correspondence and retrieval losses. Dotted boxes denote the common context; highlighted players are illustrative only and no identity or tracking information enters the model.}
    \label{fig:contextual-overview}
\end{figure*}

\subsection{Person Re-Identification}

Conventional person ReID learns identity-discriminative crop embeddings through metric learning and strong training recipes \cite{hermans2017triplet,luo2020strong}. TransReID models long-range spatial interactions with a transformer \cite{he2021transreid}, while SOLIDER learns general human-centric representations through semantic-controllable pretraining \cite{chen2023solider}. These methods provide our single-crop reference baselines. For occluded and ambiguous detections, BPBReID learns body-part representations \cite{somers2023bpbreid}, and KPR uses keypoint prompts to identify the intended person \cite{somers2024kpr}. PRTReID extends sports identity learning with team-affiliation and role-classification supervision \cite{Mansourian2023Multitask}. ContextReID instead adapts a target representation using surrounding players and observations from other frames, without requiring context identity or attribute labels at inference.

\subsection{Vision-Language Models for ReID}

CLIP-ReID adapts image--text pretraining to person retrieval through learned identity-related prompts \cite{radford2021clip,li2023clipreid}. LVLM-ReID converts a large vision-language model into a discriminative person feature extractor by generating and refining a semantic token for an individual image \cite{wang2026lvlmreid}. We similarly use a VLM to produce retrieval vectors, building on Qwen2-VL and Qwen3-VL-Embedding \cite{wang2024qwen2vl,li2026qwen3vlembedding}. Our focus is visual in-context adaptation: the trained backbone conditions each target embedding on supplied unlabeled player crops rather than encoding that crop independently. This changes the representation at inference without parameter updates.

\subsection{Temporal Evidence and Player Identity}

SportsMOT and DanceTrack expose the difficulty of associating people with similar appearances \cite{cui2023sportsmot,sun2022dancetrack}, while SoccerNet player retrieval and game-state reconstruction emphasize identity in structured sports scenes \cite{giancola2022soccernetv3,somers2024soccernetgsr}. Group-aware ReID uses graphs of neighboring group members to improve individual retrieval \cite{yan2023groupreid}. Video ReID typically aggregates target tracklets, while sports trackers use appearance for association \cite{zhang2020multicamera}. MOTIP formulates tracking as in-context ID prediction, decoding current detection IDs from historical trajectories carrying ID information \cite{gao2025motip}. ContextReID requires no target tracklet or cross-frame association: its context comprises unassociated detections, and the target identity need not occur in the shared context. It retains crop-level retrieval while using spatial and temporal evidence to adapt the embedding.

\section{Method}

\subsection{Overall Design}

ContextReID learns a retrieval representation that adapts to the visual evidence supplied at inference. Given a target player crop, detections of other players in its frame, and a small set of context frames, the model produces a normalized embedding conditioned on these observations. We refer to this process as visual in-context adaptation: the supplied context changes the target representation through attention, while the trained model parameters remain fixed. Context crops carry no identity labels and require no cross-frame association.

The design draws inspiration from in-context learning, where input examples condition a model's behavior without parameter updates \cite{brown2020language}. Here, unlabeled player observations provide visual evidence for distinguishing the target from surrounding alternatives. The model is trained to use this evidence through two complementary objectives: contextual identity matching encourages relationships among jointly encoded observations, while retrieval learning makes embeddings from separate forward passes comparable by cosine similarity. Identity annotations define supervision during training but are never supplied as model inputs.

Let $F_t=\{x_{t,1},\ldots,x_{t,N_t}\}$ contain the detected player crops at time $t$, where $x_{t,n}$ is the $n$-th crop. During training, a shared context $\mathcal{C}$ is followed by either an anchor frame $A$ or a query frame $Q$. The same backbone processes $[\mathcal{C},A]$ and $[\mathcal{C},Q]$ in separate passes, producing contextual candidate embeddings and query embeddings, respectively. Figure~\ref{fig:contextual-overview} summarizes this construction and the associated training objectives.

\subsection{Frame-Wise Contextual Input}

We represent a small set of selected frames as a multi-turn vision-language conversation. Each frame forms one user turn containing the fixed prompt \texttt{frame:} followed by its player crops. Frame boundaries preserve the grouping of observations, while the input contains no player names, identity labels, team attributes, or cross-frame associations. For $M$ selected frames, the input takes the form
\begin{equation}
 \begin{aligned}
 &[\texttt{frame:}\;x_{1,1},\ldots,x_{1,N_1}],\ldots,\\
 &[\texttt{frame:}\;x_{M,1},\ldots,x_{M,N_M}].
 \end{aligned}
\end{equation}
The model infers relationships among these observations from their visual content. Only a few frames are required; the full video is not supplied.

For each crop $x_{t,n}$, we extract the hidden state $h_{t,n}^{\mathrm{end}}$ at its final visual token and normalize it:
\begin{equation}
 z_{t,n}=\frac{h_{t,n}^{\mathrm{end}}}
 {\|h_{t,n}^{\mathrm{end}}\|_2}.
\end{equation}
This produces a crop-specific retrieval vector conditioned on the observations preceding it. Because the backbone uses causal attention, we append the anchor or query frame after the shared context prefix.

During training, a $P$--$K$ sampler selects up to $P$ players from a query frame, an anchor frame, and up to $K-2$ context frames from the same sequence. Identity annotations guide this sampling but are not included in the model input. During training, we shuffle the player-crop order within each frame so that fixed crop positions do not reveal cross-frame identity correspondences. When a selected identity is absent from another frame, its position is filled with another player from that frame, preventing fixed positions from specifying correspondence. Whenever possible, we also remove one query--anchor shared identity from the context, providing training examples in which the target must be matched without another observation of that identity in the context.

\subsection{Contextual and Retrieval Objectives}

We train the model with two complementary objectives. Contextual correspondence learning encourages the model to extract identity-discriminative evidence from the supplied observations, while retrieval learning makes embeddings produced in separate forward passes comparable for player matching. Training uses identity annotations only to form positive and negative sets outside the model. Let $s(z_i,z_j)=\tau z_i^{\top}z_j$ denote scaled cosine similarity with temperature scale $\tau$. For an embedding $z_i$, a designated positive $z_{p(i)}$, and negative index set $\mathcal{N}_i$, we use the contrastive classification loss
\begin{equation}
 \ell_i=-\log\frac{\exp(s(z_i,z_{p(i)}))}
 {\exp(s(z_i,z_{p(i)}))+\sum_{j\in\mathcal{N}_i}\exp(s(z_i,z_j))}.
\end{equation}
If an identity has multiple positive occurrences, we designate its last matching occurrence as the positive and exclude the remaining true matches from the negative set. Rows without a valid positive are excluded from the corresponding loss.

\paragraph{Contextual correspondence loss.}
This objective encourages the model to extract identity-discriminative evidence from the supplied observations. The contextual pass jointly encodes the context frames followed by the anchor frame. For every crop whose identity occurs in another frame of this pool, we retrieve the designated matching crop against all other identities in the same example. This yields
\begin{equation}
 \mathcal{L}_{\mathrm{ctx}}=\frac{1}{|\mathcal{V}_{\mathrm{ctx}}|}
 \sum_{i\in\mathcal{V}_{\mathrm{ctx}}}\ell_i,
\end{equation}
where $\mathcal{V}_{\mathrm{ctx}}$ is the set of crops with at least one cross-frame match. Players within a frame are guaranteed to be distinct and therefore provide natural hard negatives, while repeated identities across frames provide positives. Since the anchor is last, the designated last match corresponds to its anchor occurrence whenever that identity is present there.

\paragraph{Retrieval loss.}
This objective makes embeddings produced in separate forward passes comparable for player matching. The retrieval pass separately encodes the same context followed by the query frame. Only embeddings from the final query frame are supervised. Each query embedding is matched against the contextual candidate pool produced by the context-plus-anchor pass. The anchor occurrence of the same identity is the positive, other occurrences of that identity are ignored, and all different identities are negatives:
\begin{equation}
 \mathcal{L}_{\mathrm{ret}}=\frac{1}{|\mathcal{V}_{\mathrm{ret}}|}
 \sum_{i\in\mathcal{V}_{\mathrm{ret}}}\ell_i.
\end{equation}
Here, $\mathcal{V}_{\mathrm{ret}}$ denotes the set of crops in the final query frame whose identities have a valid positive in the contextual candidate pool.
This branch preserves the retrieval interface: query and candidate vectors are produced in separate forward passes and compared only by their dot product. The overall objective averages the two groups independently before combining them,
\begin{equation}
 \mathcal{L}=\mathcal{L}_{\mathrm{ret}}+\mathcal{L}_{\mathrm{ctx}},
\end{equation}
giving the two independently averaged losses equal weight regardless of the number of valid rows contributed by each.

\subsection{Inference-Time Adaptation}

At inference, ContextReID adapts a target embedding using unlabeled player detections from its own frame and a few available frames from the sequence. Context players are neither aligned nor associated across frames, and the target identity need not appear in the context. Constructing the input requires only detections and frame membership; identity annotations are used exclusively for training supervision and evaluation.

For last-position targets, we reorder the target frame separately for each crop, placing that crop after its same-frame companions. Under causal attention, its embedding can then incorporate both the preceding context frames and the other players in its own frame. The resulting representation adapts to this visual evidence without parameter updates. Placing the target first instead removes access to its same-frame companions.

A fixed context prefix can be reused across targets within a sequence. Its key--value states can be computed once and reused during per-target encoding, amortizing part of the extraction cost. With the prefix fixed, gallery embeddings can also be stored for subsequent retrieval. Changing the shared context generally requires re-encoding gallery crops to obtain embeddings conditioned on the new context. We specify the experimental frame-selection policy and examine alternative selections in Section~\ref{sec:experiments}.

\section{Experiments}
\label{sec:experiments}

\subsection{Datasets}

We evaluate on three identity-annotated multi-object tracking datasets. SportsMOT contains broadcast clips from basketball, football, and volleyball and emphasizes rapid motion, moving cameras, and visually similar teammates \cite{cui2023sportsmot}. DanceTrack contains group-dance videos in which people have highly uniform appearance but diverse and non-linear motion \cite{sun2022dancetrack}. SN-GSR-2025 is the 2025 release of the SoccerNet Game State Reconstruction benchmark; its 30-second soccer clips annotate players, goalkeepers, and referees together with persistent within-clip track identities \cite{somers2024soccernetgsr}. We convert every annotated person box into a player crop. For SN-GSR, we retain the player, goalkeeper, and referee categories.

Table~\ref{tab:dataset-statistics} summarizes the frozen training, development, and test splits. SportsMOT's labeled validation split is partitioned into 15 development and 30 test sequences, preserving source-video groups. DanceTrack's validation split is partitioned into 8 development and 17 test sequences. For SoccerNet-GSR, we use 19 official validation sequences for development and all 49 labeled official test sequences for testing; the remaining validation sequences are unused. Development data are used for checkpoint selection, separately from final testing. An identity is a distinct track within one sequence, and counts are summed over sequences rather than globally deduplicated.

\begin{table}[t]
\centering
\small
\setlength{\tabcolsep}{2.2pt}
\begin{tabular}{llrrrr}
\toprule
Dataset & Set & Seq. & Frames & IDs & Crops \\
\midrule
SportsMOT & Train & 45 & 28,574 & 639 & 312,579 \\
          & Dev & 15 & 377 & 215 & 4,175 \\
          & Test & 30 & 719 & 426 & 7,814 \\
\midrule
DanceTrack & Train & 40 & 41,796 & 419 & 348,930 \\
           & Dev & 8 & 393 & 97 & 3,378 \\
           & Test & 17 & 903 & 176 & 8,071 \\
\midrule
SN-GSR-2025 & Train & 57 & 42,750 & 1,349 & 690,624 \\
            & Dev & 19 & 570 & 436 & 9,231 \\
            & Test & 49 & 1,470 & 1,139 & 21,216 \\
\bottomrule
\end{tabular}
\caption{Dataset statistics after selecting annotated person categories. Dev and Test report retrieval inputs after approximately 1-FPS sampling; Train uses all annotated frames. IDs count sequence-local tracks present in the reported frames.}
\label{tab:dataset-statistics}
\end{table}

\subsection{Evaluation Protocol}

We construct a fixed retrieval benchmark from the held-out test sequences in Table~\ref{tab:dataset-statistics}. SportsMOT and DanceTrack use held-out portions of their labeled validation splits because their public test annotations do not contain ground-truth identities, whereas SN-GSR-2025 uses its labeled official test split. We sample annotated frames at approximately one frame per second using the native frame rate of each dataset. With a fixed random seed, one sampled frame from every sequence is selected as the query frame; every person crop in that frame forms a separate query. The query frame is excluded from the shared cross-frame context pool. Context frames are excluded from gallery ranking, leaving the remaining sampled non-query frames as retrieval candidates.

Ranking is performed independently within each sequence. A gallery crop is relevant exactly when it shares the query sequence and track identifier. We avoid pooling galleries across sequences because the datasets provide no global person identity, and distinct clips can come from the same game; treating an unannotated recurrence as a negative would therefore introduce label noise. Queries whose identity does not recur in the gallery are excluded. We rank normalized embeddings by cosine similarity and report mean average precision (mAP) and CMC Rank-1 accuracy.

\subsection{Baselines and Implementation}

We retrain three single-crop baselines on the combined training splits: SOLIDER (Swin-Base) \cite{chen2023solider}, TransReID (ViT-B/16) \cite{he2021transreid}, and R50-IBN-a \cite{luo2020strong}. All use $128\times256$ crops, sequence-local identity supervision, and a triplet objective. Unless otherwise stated, training uses $P=20$, $K=5$ crop sampling and 25,560 optimizer updates with SGD and cosine learning-rate decay; checkpoints maximize mean development mAP. We additionally evaluate the released PRTReID (HRNet-W32) checkpoint on SN-GSR-2025 \cite{Mansourian2023Multitask}.

ContextReID uses Qwen2-VL-2B-Instruct \cite{wang2024qwen2vl} or Qwen3-VL-Embedding-2B \cite{li2026qwen3vlembedding}, with $128\times256$ source crops and crop-specific \texttt{<|vision\_end|>} pooling. We fine-tune with LoRA \cite{hu2022lora} for approximately 14,100 optimizer updates, using a learning rate of $10^{-4}$ and $P=20$, $K=5$ frame sampling. Checkpoints minimize development query contrastive loss. We report update counts because the baseline and ContextReID samplers define epochs differently.

Results are averaged over three training seeds, except where noted. Additional implementation details and extended experimental results are provided in the supplementary material.

\subsection{Main Results}

\begin{table}[t]
\centering
\small
\setlength{\tabcolsep}{1.8pt}
\begin{tabular}{@{}lccrrrrrr@{}}
\toprule
& \multicolumn{2}{c}{} & \multicolumn{2}{c}{SportsMOT} & \multicolumn{2}{c}{DanceTrack} & \multicolumn{2}{c}{SN-GSR} \\
\cmidrule(lr){4-5}\cmidrule(lr){6-7}\cmidrule(lr){8-9}
Method & CF & TP & mAP & R-1 & mAP & R-1 & mAP & R-1 \\
\midrule
PRTReID & -- & -- & -- & -- & -- & -- & 69.96 & 85.91 \\
R50-IBN-a & -- & -- & 68.35 & 88.52 & 25.56 & 59.14 & 76.64 & 89.87 \\
TransReID & -- & -- & 70.07 & 88.52 & 23.91 & 49.03 & 79.26 & 91.34 \\
SOLIDER$^{\dagger}$ & -- & -- & 81.79 & 94.76 & 26.84 & 56.13 & 82.41 & 92.75 \\
\midrule
\multicolumn{9}{c}{Qwen2-VL-2B-Instruct} \\
\midrule
Zero-shot & -- & -- & 23.66 & 43.11 & 16.83 & 34.19 & 24.24 & 45.30 \\
\hdashline[0.5pt/2pt]
\multirow{5}{*}{ContextReID} & 0 & F & 79.59 & 92.02 & 33.15 & 57.63 & 78.14 & 89.00 \\
 & 0 & L & 84.55 & 94.01 & 37.27 & 59.78 & 81.47 & 91.16 \\
 & 1 & L & 86.51 & 95.11 & 39.02 & 60.86 & 83.40 & 91.57 \\
 & 2 & L & 87.19 & 95.11 & 39.29 & 60.43 & 83.70 & 91.25 \\
 & 3 & L & 87.32 & 94.81 & 39.05 & 58.49 & 83.47 & 90.42 \\
\midrule
\multicolumn{9}{c}{Qwen3-VL-Embedding-2B} \\
\midrule
Zero-shot & -- & -- & 39.85 & 60.48 & 23.79 & 43.23 & 39.97 & 65.06 \\
\hdashline[0.5pt/2pt]
\multirow{5}{*}{ContextReID} & 0 & F & 82.56 & 93.51 & 36.52 & \underline{63.23} & 80.04 & 90.93 \\
 & 0 & L & 86.10 & 94.41 & 40.99 & \textbf{64.30} & 83.16 & 92.40 \\
 & 1 & L & 89.03 & 96.01 & 43.39 & 62.80 & 85.93 & \textbf{93.37} \\
 & 2 & L & \underline{89.92} & \textbf{96.41} & \textbf{44.04} & 60.86 & \underline{86.60} & \underline{93.00} \\
 & 3 & L & \textbf{90.67} & \underline{96.21} & \underline{43.57} & 58.49 & \textbf{86.64} & \textbf{93.37} \\
\bottomrule
\end{tabular}
\caption{Player retrieval results (\%). R-1 denotes CMC Rank-1. Models report means over three training seeds, except SOLIDER$^{\dagger}$, which averages two seeds; one seed was excluded due to NaN loss. Full mean $\pm$ standard deviation results are provided in the supplementary material. Best values are bold and second-best values are underlined.}
\label{tab:main-results}
\end{table}

\begin{table*}[t]
\centering
\small
\setlength{\tabcolsep}{3.5pt}
\begin{tabular}{@{}lccrrrr||rrrr||rrrr@{}}
\toprule
\multicolumn{3}{c}{Source / Target} & \multicolumn{2}{c}{B + V} & \multicolumn{2}{c||}{S} & \multicolumn{2}{c}{B + S} & \multicolumn{2}{c||}{V} & \multicolumn{2}{c}{S + V} & \multicolumn{2}{c}{B} \\
\cmidrule(lr){4-5}\cmidrule(lr){6-7}\cmidrule(lr){8-9}\cmidrule(lr){10-11}\cmidrule(lr){12-13}\cmidrule(lr){14-15}
Method & CF & TP & mAP & R-1 & mAP & R-1 & mAP & R-1 & mAP & R-1 & mAP & R-1 & mAP & R-1 \\
\midrule
SOLIDER (Swin-B) & -- & -- & 82.82 & 94.05 & 55.00 & 77.96 & 75.02 & \textbf{92.69} & 76.38 & 90.72 & 78.38 & \underline{91.38} & 63.06 & 90.82 \\
\midrule
\multirow{5}{*}{\shortstack{ContextReID\\w/\\Qwen3-VL-2B}} & 0 & F & 84.01 & 93.11 & 60.30 & 80.99 & 77.81 & 90.41 & 76.35 & 88.70 & 76.09 & 89.12 & 76.38 & 92.86 \\
 & 0 & L & 86.36 & 94.84 & 61.14 & 78.51 & 79.96 & 91.02 & 78.47 & \textbf{91.88} & 78.44 & \underline{91.38} & 79.14 & \textbf{95.24} \\
 & 1 & L & 89.32 & 95.93 & 63.24 & 78.79 & 82.54 & 91.17 & 79.80 & \underline{91.59} & 80.32 & 89.97 & 83.47 & \textbf{95.24} \\
 & 2 & L & \underline{90.39} & \underline{96.09} & \underline{64.29} & \underline{83.75} & \underline{83.77} & \underline{91.93} & \underline{81.21} & 91.01 & \underline{81.40} & \textbf{91.53} & \underline{84.75} & \underline{93.88} \\
 & 3 & L & \textbf{91.50} & \textbf{96.24} & \textbf{64.39} & \textbf{84.30} & \textbf{84.07} & 91.78 & \textbf{83.39} & 90.43 & \textbf{82.59} & 91.24 & \textbf{85.60} & \underline{93.88} \\
\bottomrule
\end{tabular}
\caption{Cross-sport transfer on SportsMOT (\%), averaged over three training seeds. B, V, and S denote basketball, volleyball, and soccer (the dataset's football category), respectively. Each model is trained on the two source sports and evaluated on both the held-out target sport and independent test sequences from the source sports. Best and second-best values within each metric column are bold and underlined, respectively; ties receive the same emphasis.}
\label{tab:cross-sport-transfer}
\end{table*}

Table~\ref{tab:main-results} compares ContextReID with conventional single-crop ReID models and summarizes the context-count variants used in our ablation study. CF denotes the number of inference-time context frames and TP denotes the target-crop position, with F and L indicating first and last, respectively. Because the VLM uses causal attention, TP controls whether the target embedding can incorporate other players from its own frame. With TP=F, the target crop precedes its same-frame teammates and therefore cannot attend to them; with TP=L, it follows the other crops and can use their appearance as same-frame context. For the main results, we fix the context pool to the first three sampled non-query frames of each sequence. CF=1, 2, and 3 use the first one, two, and three frames of this pool, respectively; CF=0 uses no cross-frame context. All crops from the three-frame pool are excluded from gallery ranking for every method and configuration, including the single-crop baselines. This shared protocol prevents context crops from also serving as retrieval candidates and keeps every method on the same query--gallery split.

The released zero-shot checkpoints provide a reference for the effect of task-specific training. Both Qwen2-VL-2B-Instruct and Qwen3-VL-Embedding-2B are substantially below their fine-tuned ContextReID counterparts under isolated-crop inference indicating the importance of adapting the pretrained representations to player retrieval.

The conventional baselines serve as reference points for single-crop retrieval rather than parameter-matched comparisons: their architectures, parameter counts, and pretraining differ from those of ContextReID. Without in-context adaptation (CF=0, TP=F), each target is represented as an isolated crop. Both ContextReID variants achieve broadly comparable retrieval performance to SOLIDER, although differences vary across datasets and metrics. For example, SportsMOT mAP is 79.59\% for Qwen-2 and 82.56\% for Qwen-3, compared with 81.79\% for SOLIDER. On DanceTrack, both variants already achieve higher mAP in this isolated-crop setting. Thus, the larger backbones do not uniformly yield large gains under isolated-crop inference. More directly, comparing each ContextReID checkpoint with and without context holds its parameters and training fixed, allowing us to assess the additional benefit of in-context adaptation independently of model size.

Both Qwen-2 and Qwen-3 benefit from in-context adaptation. With the fixed CF=1, TP=L setting, mAP improves over their respective isolated-crop results on all three datasets: by 6.92, 5.87, and 5.26 percentage points for Qwen-2, and by 6.47, 6.87, and 5.89 points for Qwen-3 on SportsMOT, DanceTrack, and SN-GSR, respectively. These gains arise without parameter updates or identity labels for the supplied context at inference. Additional context frames further improve mAP in several settings, while Rank-1 gains are less uniform. The results support the value of in-context adaptation across both backbones.

\subsection{Cross-Sport Transfer}

We evaluate cross-sport transfer with a leave-one-sport-out protocol on SportsMOT. For each target sport, we train the Qwen-3 variant and SOLIDER, the strongest baseline, on the other two sports and select checkpoints using development data from the source sports only. The target sport is excluded from task-specific training and checkpoint selection. At inference, the model receives unlabeled visual context from the target sequence without parameter updates. Table~\ref{tab:cross-sport-transfer} reports means over three training seeds; full mean $\pm$ standard deviation results and transfer-specific training settings are provided in the supplementary material.

Even without in-context adaptation (CF=0, TP=F), Qwen-3 shows a transfer advantage over the Swin baseline when soccer or basketball is held out. On soccer, it achieves 60.30\% mAP and 80.99\% Rank-1, compared with 55.00\% and 77.96\% for Swin; on basketball, it achieves 76.38\% mAP and 92.86\% Rank-1, compared with 63.06\% and 90.82\%. On held-out volleyball, isolated-crop mAP is similar (76.35\% versus 76.38\%), while Rank-1 is lower. These results indicate that the backbone already provides useful transfer capabilities before contextual adaptation.

In-context adaptation further improves retrieval, increasing mAP on all three held-out sports. With CF=1, TP=L, the gains over isolated-crop inference are 2.94 points on soccer, 3.45 on volleyball, and 7.09 on basketball. With three context frames, these gains reach 4.09, 7.04, and 9.22 points, respectively. Rank-1 gains vary across context settings, although each held-out sport benefits from at least one contextual configuration. These within-checkpoint comparisons show that unlabeled visual context can improve retrieval beyond the backbone's existing transfer capability, without additional parameter updates.

\subsection{Ablation Studies}

We use the ContextReID variants in Table~\ref{tab:main-results} to study the effects of same-frame and cross-frame context. The CF=0, TP=F configuration represents the performance of the Qwen-based model under a conventional ReID setting, where each target is embedded without cross-frame or same-frame player context. Moving the target to the last position at CF=0 supplies same-frame context and improves both metrics on all three datasets for both backbones.

For Qwen-3, mAP peaks at three context frames on SportsMOT and SN-GSR-2025 and at two on DanceTrack. Rank-1 peaks at two on SportsMOT, at zero with the target last on DanceTrack, and ties at one and three on SN-GSR-2025. These comparisons measure the effect of inference-time context, which support that both same-frame player context and observations from additional frames contribute to the learned representation, although the optimal number of context frames varies by metric and dataset.

\paragraph{Context-frame selection.}
We examine sensitivity to the supplied context using the Qwen-3 variant with CF=1, TP=L. We select the first, middle, or last sampled non-query frame, retaining the same query set and a common gallery that excludes all three candidate context frames. Table~\ref{tab:context-selection} reports three-seed mean mAP and sample standard deviation. The preferred selection varies across datasets: the middle frame yields the highest mAP on SportsMOT and DanceTrack, while the first frame performs best on SN-GSR. Nevertheless, the spread across selection means is limited to 1.86, 0.41, and 0.77 points, respectively, indicating stable retrieval performance across the tested choices. Together with the contextual gains in the main results, this suggests that useful adaptation is not restricted to the first-frame policy.

\begin{table}[t]
\centering
\small
\setlength{\tabcolsep}{3pt}
\begin{tabular}{@{}lccc@{}}
\toprule
Context & SportsMOT & DanceTrack & SN-GSR \\
\midrule
First & $88.93 \pm 0.11$ & $43.27 \pm 0.19$ & $86.03 \pm 0.38$ \\
Middle & $90.79 \pm 0.09$ & $43.68 \pm 0.04$ & $85.26 \pm 0.31$ \\
Last & $90.26 \pm 0.24$ & $43.39 \pm 0.22$ & $85.46 \pm 0.05$ \\
\midrule
Overall & $89.99 \pm 0.13$ & $43.45 \pm 0.14$ & $85.58 \pm 0.24$ \\
\bottomrule
\end{tabular}
\caption{Context-frame selection with Qwen-3 (CF=1, TP=L): mean $\pm$ sample standard deviation of mAP (\%) over three seeds. Overall mean averages the three selection policies.}
\label{tab:context-selection}
\end{table}

\begin{figure*}[t]
\begin{minipage}[t]{0.48\textwidth}
\vspace{0pt}
    \centering
    \includegraphics[width=\columnwidth]{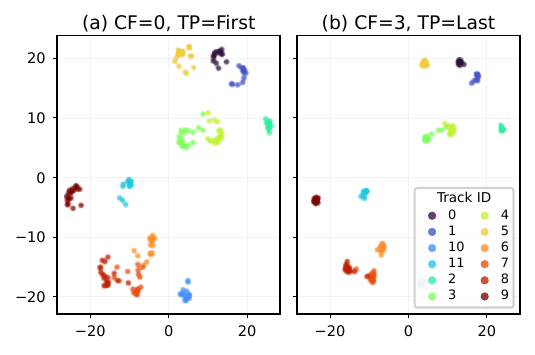}
    \caption{Joint t-SNE visualization for the same player crops from one SportsMOT sequence. Both configurations are projected together using cosine distance; colors and numeric labels denote ground-truth track identities. Compared with isolated-crop embeddings (left), contextualized embeddings using three context frames and a last-position target (right) form more compact within-identity groups.}
    \label{fig:tsne-case-study}
\end{minipage}\hfill
\begin{minipage}[t]{0.48\textwidth}
\vspace{0pt}
    \centering
    \includegraphics[width=\columnwidth]{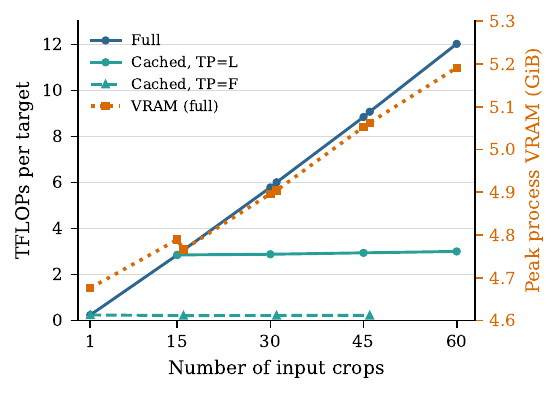}
\caption{Qwen-3 computational footprint. Left axis: full versus cached computation per target, excluding one-time prefix construction for cached curves. Right axis: full-forward BF16/SDPA peak process VRAM. Full and cached memory peaks are similar, so only full-forward results are shown.}
    \label{fig:computational-footprint}
\end{minipage}
\end{figure*}

\paragraph{Context-content control.}
We test whether gains arise merely from additional visual input by replacing the shared context with seeded random frames from an unrelated dataset. SportsMOT and SN-GSR-2025 use DanceTrack donors, while DanceTrack uses SN-GSR-2025 donors. All settings use the Qwen-3 variant with TP=F. Table~\ref{tab:context-control} compares three-seed mean mAP against the CF=0, TP=F reference from Table~\ref{tab:main-results}. Unrelated context decreases mAP on SportsMOT and DanceTrack; on SN-GSR, the largest gain is 0.52 points. These changes are small relative to the gains from sequence context in the main results, suggesting that additional visual input alone is insufficient to explain those gains.

\begin{table}[t]
\centering
\small
\setlength{\tabcolsep}{3.5pt}
\begin{tabular}{@{}crrr@{}}
\toprule
& \multicolumn{3}{c}{mAP / $\Delta$mAP} \\
\cmidrule(lr){2-4}
CF & SportsMOT & DanceTrack & SN-GSR \\
\midrule
0 & 82.56 & 36.52 & 80.04 \\
\midrule
1 & -0.17 & -0.04 & +0.52 \\
2 & -0.34 & -0.15 & +0.42 \\
3 & -0.65 & -0.41 & +0.01 \\
\bottomrule
\end{tabular}
\caption{Qwen-3 context-content control with TP=F, averaged over three seeds. The first row reports absolute CF=0, TP=F mAP (\%) from Table~\ref{tab:main-results}; subsequent rows report signed mAP percentage-point changes relative to this reference. Unrelated context comprises seeded random frames drawn from a different dataset.}
\label{tab:context-control}
\end{table}

\begin{table}[t]
\centering
\small
\setlength{\tabcolsep}{3.5pt}
\begin{tabular}{@{}ccrrr@{}}
\toprule
& & \multicolumn{3}{c}{mAP / $\Delta$mAP} \\
\cmidrule(lr){3-5}
CF & TP & SportsMOT & DanceTrack & SN-GSR \\
\midrule
0 & F & 89.21 & 43.94 & 83.83 \\
\midrule
0 & L & +1.89 & +3.77 & +3.23 \\
1 & L & +2.37 & +4.71 & +3.02 \\
2 & L & +3.00 & +3.26 & +2.14 \\
3 & L & +1.26 & +2.72 & +1.40 \\
\bottomrule
\end{tabular}
\caption{Qwen-3 target-absence control on a fixed one-identity-per-sequence query subset, averaged over three seeds. The first row reports absolute single-crop mAP (\%); subsequent rows report signed mAP percentage-point changes relative to it. The designated query identity is removed from shared cross-frame context.}
\label{tab:target-absence-control}
\end{table}

\paragraph{Target-absence control.}
We test whether contextual gains require the query identity to appear in the shared cross-frame context. We retain one query identity per sequence (30 SportsMOT, 17 DanceTrack, and 49 SN-GSR-2025 queries) and remove it from the shared context for both query and gallery encoding. Own-frame context and gallery membership remain unchanged. Table~\ref{tab:target-absence-control} reports three-seed Qwen-3 means. Every contextual setting improves mAP over isolated crops. Relative to same-frame context alone (CF=0, TP=L), the best additional gain is 1.11 points on SportsMOT (CF=2) and 0.94 on DanceTrack (CF=1). On SN-GSR-2025, same-frame context alone gives the highest mAP. These results support adaptation from surrounding players.

\subsection{Embedding Case Study}

Figure~\ref{fig:tsne-case-study} qualitatively compares the isolated-crop and contextualized embedding spaces for one SportsMOT sequence. We jointly fit a single t-SNE projection to the embeddings from both configurations, using the same set of crops in each, and color points by ground-truth track identity. Under CF=0, TP=F, several identities exhibit visibly dispersed embeddings across their observations. With CF=3, TP=L, observations of the same identity form more compact groups while remaining separated from other identities. Although t-SNE is only a qualitative visualization, this example illustrates how contextualization can reduce variation across repeated observations and produce more coherent identity structure.

\subsection{Computational Analysis}

We profile the Qwen3-VL-Embedding-2B variant in BF16 using SportsMOT inputs, grouping complete forwards by the total number of player crops. FLOPs are counted with eager attention, treating a multiply--add as two operations; runtime peak process VRAM is measured separately with SDPA and NVML. Each input uses 32 visual tokens per crop.

Figure~\ref{fig:computational-footprint} compares full recomputation with prefix-cache reuse on fixed examples containing 15 crops per frame. With one, two, and three context frames and TP=L, full forwards require 5.78, 8.84, and 12.02 TFLOPs, whereas encoding a target with an existing prefix cache requires 2.88, 2.94, and 3.00 TFLOPs. These cached costs exclude the one-time prefix construction. TP=F further reduces computation because only the target crop is processed after the prefix. The right axis shows full-forward process-memory peaks, increasing from 4.68~GiB for one crop to 5.19~GiB for 60 crops. Matched full and cached probing gives similar process-memory peaks across the tested settings, with absolute differences below 0.05~GiB; we therefore display only the full-forward measurements. Cache reuse primarily reduces repeated computation rather than peak process VRAM in these measurements.

\section{Conclusion}

We presented ContextReID, a framework for visual in-context adaptation in player re-identification. By conditioning a target embedding on co-occurring players and observations from a few additional frames, the trained model uses spatial and temporal evidence without parameter updates, context identity labels, or cross-frame identity associations at inference. Experiments across two vision-language backbones and three benchmarks demonstrate retrieval improvements beyond the same models' isolated-crop representations. Cross-sport experiments further show that contextual adaptation improves mAP on sports excluded from task-specific training. Controlled analyses support the value of relevant visual context, including when the designated query identity is absent from the shared context. These findings highlight the empirical value of adapting player representations to the visual evidence available at inference.

\section*{LLM Usage Statement}
OpenAI Codex assisted with manuscript drafting and revision, bibliographic
verification, \LaTeX{} formatting, and preparation of figures and tables from
experimental outputs. It also assisted with code development, dataset
preparation, experiment execution, and analysis of evaluation results.
The evaluated ContextReID variants use Qwen2-VL-2B-Instruct and
Qwen3-VL-Embedding-2B as described in the method and experiments.

{\small
\bibliographystyle{ieeenat_fullname}
\bibliography{cvpr27}
}

\clearpage
\appendix
\setcounter{table}{0}
\setcounter{figure}{0}
\renewcommand{\thetable}{S\arabic{table}}
\renewcommand{\thefigure}{S\arabic{figure}}
\renewcommand{\theHtable}{supp.\arabic{table}}
\renewcommand{\theHfigure}{supp.\arabic{figure}}
\maketitlesupplementary

\section{Detailed Implementation Setup}

This section expands the baseline and ContextReID implementation details described in Section 4.3 of the main paper. Unless otherwise stated, the following settings apply to the main three-dataset comparison; the cross-sport baseline uses the separate settings described below.

\subsection{Models and Initialization}
We retrain SOLIDER with Swin-Base \cite{chen2023solider}, TransReID with ViT-B/16 \cite{he2021transreid}, and the ResNet50-IBN-a strong baseline \cite{luo2020strong}. SOLIDER starts from its released human-centric pretrained weights, while TransReID and R50-IBN-a start from ImageNet-pretrained weights. All models receive individual player crops resized to width 128 and height 256 pixels. Training uses random horizontal flipping, padding followed by random cropping, normalization, and random erasing. Evaluation uses resizing and normalization without stochastic augmentation.

\subsection{Sampling and Supervision}
All retrained baselines use the combined training sequences of SportsMOT, DanceTrack, and SoccerNet-GSR under the frozen development/test partition in the main paper. We construct sequence-restricted identity batches with up to $P=20$ identities and $K=5$ crops per identity, giving at most 100 crops per batch. Track identities are scoped to individual sequences. We train with a triplet objective and omit identity classification to avoid additional assumptions about identities across clips. Training uses BF16 mixed precision; the TransReID and SOLIDER runs use the numerically stable softplus formulation of the soft-margin triplet loss, while R50-IBN-a uses the original loss implementation.

A baseline epoch draws $K$ crops per identity rather than traversing every annotated crop. The main training schedule contains 213 batches per epoch and 120 epochs, corresponding to 25,560 optimizer-update steps. These epochs are therefore not directly comparable to ContextReID's frame-based sampling epochs.

\subsection{Optimization and Checkpoint Selection}
We use SGD with weight decay $10^{-4}$ and a cosine learning-rate schedule. The base learning rates are $2\times10^{-4}$ for SOLIDER, $8\times10^{-3}$ for TransReID, and $2\times10^{-3}$ for R50-IBN-a. The first 20 epochs use warmup starting at 1\% of the base rate; the cosine scheduler has a minimum rate of 0.2\% of the base rate.

Development retrieval is evaluated every ten epochs. We select the checkpoint maximizing the unweighted mean development mAP across the three datasets; test scores are not used for selection. Development retrieval uses the sampled sequence-local query/gallery protocol described in the main paper.

Runs use training seeds 42, 43, and 44 with fixed development/test selections. SOLIDER's main-table mean uses seeds 42 and 44 because seed 43 encountered NaN loss. The supplementary numerical results should be interpreted with this two-run exception.

\subsection{Released PRTReID Reference}
For PRTReID \cite{Mansourian2023Multitask}, we use the HRNet-W32 checkpoint distributed with the SoccerNet Game State Reconstruction baseline, without retraining. We report this reference on SoccerNet-GSR only. Its saved crop embeddings are evaluated under the current sequence-restricted test protocol and shared non-query context-frame gallery exclusions. As a fixed released checkpoint, PRTReID is not included in the three-seed training protocol.

\subsection{Cross-Sport Transfer Baseline}
The leave-one-sport-out study retrains SOLIDER on the two source sports using BF16, the softplus triplet loss, and $P=20$, $K=5$ sampling. These runs use a reduced base learning rate of $10^{-5}$, a 600-epoch maximum, and early stopping after 100 epochs without improvement in source-development mAP. Each seed starts from the released pretrained weights. Checkpoints are selected exclusively using source-development mAP and then evaluated on both source-sport test sequences and the held-out sport. Source-test aggregates weight each sport's metric by its valid-query count within a run before averaging across seeds 42, 43, and 44.

\begin{table*}[t]
\centering
\small
\setlength{\tabcolsep}{3pt}
\begin{tabular}{@{}lcccccccc@{}}
\toprule
& & & \multicolumn{2}{c}{SportsMOT} & \multicolumn{2}{c}{DanceTrack} & \multicolumn{2}{c}{SoccerNet-GSR} \\
\cmidrule(lr){4-5}\cmidrule(lr){6-7}\cmidrule(lr){8-9}
Method & CF & TP & \multicolumn{1}{c}{mAP} & \multicolumn{1}{c}{R-1} & \multicolumn{1}{c}{mAP} & \multicolumn{1}{c}{R-1} & \multicolumn{1}{c}{mAP} & \multicolumn{1}{c}{R-1} \\
\midrule
PRTReID & -- & -- & -- & -- & -- & -- & 69.96 & 85.91 \\
R50-IBN-a & -- & -- & $68.35 \pm 0.45$ & $88.52 \pm 1.05$ & $25.56 \pm 0.56$ & $59.14 \pm 0.37$ & $76.64 \pm 0.52$ & $89.87 \pm 0.79$ \\
TransReID & -- & -- & $70.07 \pm 1.07$ & $88.52 \pm 0.86$ & $23.91 \pm 0.89$ & $49.03 \pm 1.12$ & $79.26 \pm 0.54$ & $91.34 \pm 0.52$ \\
SOLIDER$^{\dagger}$ & -- & -- & $81.79 \pm 0.15$ & $94.76 \pm 0.21$ & $26.84 \pm 0.76$ & $56.13 \pm 2.74$ & $82.41 \pm 0.23$ & $92.75 \pm 0.68$ \\
\midrule
\multicolumn{9}{c}{Qwen2-VL-2B-Instruct} \\
\midrule
Zero-shot & -- & -- & 23.66 & 43.11 & 16.83 & 34.19 & 24.24 & 45.30 \\
ContextReID & 0 & F & $79.59 \pm 0.56$ & $92.02 \pm 0.69$ & $33.15 \pm 1.09$ & $57.63 \pm 0.99$ & $78.14 \pm 0.33$ & $89.00 \pm 0.35$ \\
ContextReID & 0 & L & $84.55 \pm 0.36$ & $94.01 \pm 1.04$ & $37.27 \pm 0.66$ & $59.78 \pm 3.31$ & $81.47 \pm 0.29$ & $91.16 \pm 0.50$ \\
ContextReID & 1 & L & $86.51 \pm 0.19$ & $95.11 \pm 0.35$ & $39.02 \pm 0.42$ & $60.86 \pm 1.97$ & $83.40 \pm 0.32$ & $91.57 \pm 0.77$ \\
ContextReID & 2 & L & $87.19 \pm 0.10$ & $95.11 \pm 0.62$ & $39.29 \pm 0.25$ & $60.43 \pm 2.98$ & $83.70 \pm 0.20$ & $91.25 \pm 0.83$ \\
ContextReID & 3 & L & $87.32 \pm 0.19$ & $94.81 \pm 0.91$ & $39.05 \pm 0.43$ & $58.49 \pm 2.44$ & $83.47 \pm 0.16$ & $90.42 \pm 0.81$ \\
\midrule
\multicolumn{9}{c}{Qwen3-VL-Embedding-2B} \\
\midrule
Zero-shot & -- & -- & 39.85 & 60.48 & 23.79 & 43.23 & 39.97 & 65.06 \\
ContextReID & 0 & F & $82.56 \pm 0.14$ & $93.51 \pm 1.13$ & $36.52 \pm 0.06$ & $63.23 \pm 1.71$ & $80.04 \pm 0.21$ & $90.93 \pm 0.49$ \\
ContextReID & 0 & L & $86.10 \pm 0.20$ & $94.41 \pm 0.62$ & $40.99 \pm 0.23$ & $64.30 \pm 1.34$ & $83.16 \pm 0.29$ & $92.40 \pm 0.63$ \\
ContextReID & 1 & L & $89.03 \pm 0.10$ & $96.01 \pm 0.75$ & $43.39 \pm 0.08$ & $62.80 \pm 1.49$ & $85.93 \pm 0.39$ & $93.37 \pm 0.24$ \\
ContextReID & 2 & L & $89.92 \pm 0.31$ & $96.41 \pm 0.60$ & $44.04 \pm 0.65$ & $60.86 \pm 1.34$ & $86.60 \pm 0.34$ & $93.00 \pm 0.68$ \\
ContextReID & 3 & L & $90.67 \pm 0.11$ & $96.21 \pm 0.62$ & $43.57 \pm 0.40$ & $58.49 \pm 1.34$ & $86.64 \pm 0.23$ & $93.37 \pm 0.77$ \\
\bottomrule
\end{tabular}
\caption{Full main results (\%): mean $\pm$ sample standard deviation across training seeds. $\dagger$ SOLIDER averages two seeds; all other retrained models average three. Fixed released checkpoints are shown without a standard deviation. CF and TP follow the definitions in the main paper.}
\label{tab:full-main}
\end{table*}

\subsection{ContextReID Configuration}
We initialize ContextReID from Qwen2-VL-2B-Instruct \cite{wang2024qwen2vl} or Qwen3-VL-Embedding-2B \cite{li2026qwen3vlembedding}. Both variants receive source crops resized to $128\times256$ pixels and pool the hidden state at each crop-specific \texttt{<|vision\_end|>} token, followed by $\ell_2$ normalization. Qwen3-VL uses $16\times16$ patches with $2\times2$ spatial pooling, yielding 32 visual tokens per crop. Qwen2-VL uses $14\times14$ patches with $2\times2$ pooling; its native processor maps the source crops to $140\times252$ pixels, yielding 45 visual tokens.

We fine-tune with LoRA \cite{hu2022lora}, using rank 8, scaling factor 8, and dropout 0.1 on the applicable \texttt{q\_proj}, \texttt{v\_proj}, and \texttt{qkv} modules. Training uses BF16 precision on two NVIDIA RTX A6000 GPUs, 8-bit PagedAdamW, a learning rate of $10^{-4}$, and a linear schedule with 10\% warmup. The per-device batch size is one example, with eight gradient-accumulation steps. The training budget is approximately 14,100 optimizer updates, and checkpoints are selected by the lowest development query contrastive loss rather than test retrieval performance.

Each training example samples up to $P=20$ players and $K=5$ frames: one query frame, one anchor frame, and up to three context frames. We shuffle crop order within frames during training to prevent order from encoding identity associations. The contextual-matching and retrieval losses use cosine-similarity scaling $\tau=20$ and are summed with equal weight. Both variants use the same training splits and sequence-local identity supervision as the retrained baselines, with seeds 42, 43, and 44. Because the identity-balanced baseline sampler and multi-frame ContextReID sampler define epochs differently, we compare training budgets in optimizer updates rather than epochs.

\section{Full Main Results}
Table~\ref{tab:full-main} expands the main-paper comparison with sample standard deviations ($n-1$ denominator). Fine-tuned results use training seeds 42, 43, and 44, except SOLIDER, which uses seeds 42 and 44 after the NaN-loss exclusion. The same frozen evaluation episodes are used for all runs. Released zero-shot checkpoints and PRTReID have no training-seed variation; we report their fixed-checkpoint scores without a standard deviation.

\section{Full Cross-Sport Transfer Results}
Tables~\ref{tab:full-transfer-0}--\ref{tab:full-transfer-2} report three-seed means and sample standard deviations for the source-test and held-out-target evaluations, covering B + V $\rightarrow$ S, B + S $\rightarrow$ V, and S + V $\rightarrow$ B, respectively. All models are selected using source-development data only. Source-test metrics are first weighted by valid-query counts across the two source sports within each seed; standard deviations are then computed across seeds. B, V, and S denote basketball, volleyball, and soccer, respectively. The training and test partitions and selected checkpoints match the main-paper transfer table.

\begin{table}[t]
\centering
\footnotesize
\setlength{\tabcolsep}{1pt}
\begin{tabular}{@{}lccc|c||c|c@{}}
\toprule
\multicolumn{7}{c}{B + V $\rightarrow$ S} \\
& & & \multicolumn{2}{c||}{Source} & \multicolumn{2}{c}{Target} \\
\cmidrule(lr){4-5}\cmidrule(lr){6-7}
Method & CF & TP & mAP & R-1 & mAP & R-1 \\
\midrule
Swin-B & -- & -- & $82.82 \pm 1.90$ & $94.05 \pm 1.43$ & $55.00 \pm 1.50$ & $77.96 \pm 1.72$ \\
Ours & 0 & F & $84.01 \pm 0.71$ & $93.11 \pm 0.27$ & $60.30 \pm 0.83$ & $80.99 \pm 3.60$ \\
Ours & 0 & L & $86.36 \pm 0.46$ & $94.84 \pm 0.00$ & $61.14 \pm 0.93$ & $78.51 \pm 0.83$ \\
Ours & 1 & L & $89.32 \pm 0.45$ & $95.93 \pm 0.72$ & $63.24 \pm 1.61$ & $78.79 \pm 0.95$ \\
Ours & 2 & L & $90.39 \pm 0.53$ & $96.09 \pm 0.27$ & $64.29 \pm 0.50$ & $83.75 \pm 1.26$ \\
Ours & 3 & L & $91.50 \pm 0.43$ & $96.24 \pm 0.47$ & $64.39 \pm 0.29$ & $84.30 \pm 1.65$ \\
\bottomrule
\end{tabular}
\caption{B + V $\rightarrow$ S transfer: mean $\pm$ sample standard deviation (\%) across three seeds. Ours denotes ContextReID with Qwen3-VL-Embedding-2B. Source scores are query-weighted across the two training sports' test sets.}
\label{tab:full-transfer-0}
\end{table}

\begin{table}[t]
\centering
\footnotesize
\setlength{\tabcolsep}{1pt}
\begin{tabular}{@{}lccc|c||c|c@{}}
\toprule
\multicolumn{7}{c}{B + S $\rightarrow$ V} \\
& & & \multicolumn{2}{c||}{Source} & \multicolumn{2}{c}{Target} \\
\cmidrule(lr){4-5}\cmidrule(lr){6-7}
Method & CF & TP & mAP & R-1 & mAP & R-1 \\
\midrule
Swin-B & -- & -- & $75.02 \pm 1.41$ & $92.69 \pm 1.21$ & $76.38 \pm 1.18$ & $90.72 \pm 1.00$ \\
Ours & 0 & F & $77.81 \pm 0.32$ & $90.41 \pm 0.91$ & $76.35 \pm 2.65$ & $88.70 \pm 2.30$ \\
Ours & 0 & L & $79.96 \pm 0.61$ & $91.02 \pm 3.04$ & $78.47 \pm 1.70$ & $91.88 \pm 2.80$ \\
Ours & 1 & L & $82.54 \pm 0.43$ & $91.17 \pm 1.73$ & $79.80 \pm 1.43$ & $91.59 \pm 1.00$ \\
Ours & 2 & L & $83.77 \pm 0.24$ & $91.93 \pm 0.95$ & $81.21 \pm 1.43$ & $91.01 \pm 1.33$ \\
Ours & 3 & L & $84.07 \pm 0.08$ & $91.78 \pm 0.46$ & $83.39 \pm 1.20$ & $90.43 \pm 0.87$ \\
\bottomrule
\end{tabular}
\caption{B + S $\rightarrow$ V transfer: mean $\pm$ sample standard deviation (\%) across three seeds. Ours denotes ContextReID with Qwen3-VL-Embedding-2B. Source scores are query-weighted across the two training sports' test sets.}
\label{tab:full-transfer-1}
\end{table}

\begin{table}[t]
\centering
\footnotesize
\setlength{\tabcolsep}{1pt}
\begin{tabular}{@{}lccc|c||c|c@{}}
\toprule
\multicolumn{7}{c}{S + V $\rightarrow$ B} \\
& & & \multicolumn{2}{c||}{Source} & \multicolumn{2}{c}{Target} \\
\cmidrule(lr){4-5}\cmidrule(lr){6-7}
Method & CF & TP & mAP & R-1 & mAP & R-1 \\
\midrule
Swin-B & -- & -- & $78.38 \pm 0.95$ & $91.38 \pm 0.49$ & $63.06 \pm 3.50$ & $90.82 \pm 2.70$ \\
Ours & 0 & F & $76.09 \pm 0.65$ & $89.12 \pm 0.49$ & $76.38 \pm 2.16$ & $92.86 \pm 3.68$ \\
Ours & 0 & L & $78.44 \pm 1.00$ & $91.38 \pm 0.88$ & $79.14 \pm 1.10$ & $95.24 \pm 0.59$ \\
Ours & 1 & L & $80.32 \pm 0.71$ & $89.97 \pm 0.24$ & $83.47 \pm 1.16$ & $95.24 \pm 1.18$ \\
Ours & 2 & L & $81.40 \pm 0.88$ & $91.53 \pm 1.53$ & $84.75 \pm 0.70$ & $93.88 \pm 1.02$ \\
Ours & 3 & L & $82.59 \pm 0.64$ & $91.24 \pm 0.49$ & $85.60 \pm 0.83$ & $93.88 \pm 1.77$ \\
\bottomrule
\end{tabular}
\caption{S + V $\rightarrow$ B transfer: mean $\pm$ sample standard deviation (\%) across three seeds. Ours denotes ContextReID with Qwen3-VL-Embedding-2B. Source scores are query-weighted across the two training sports' test sets.}
\label{tab:full-transfer-2}
\end{table}

\section{Input Format and Crop-Order Ablations}
\subsection{Training Input Format}
We compare frame-wise conversations with a flat/shuffled input using Qwen3-VL-Embedding-2B. Both variants use the same pretrained initialization, training splits, $P=20$, $K=5$ sampling, source-crop resolution, LoRA settings, optimizer schedule, and development-query-loss checkpoint selection. We train each variant with seeds 42, 43, and 44 and evaluate on the same frozen test episodes and corrected shared non-query support protocol.

During training, the flat/shuffled variant removes the \texttt{frame:} markers and frame-turn boundaries and places context crops in one user turn with their order shuffled across frames. Anchor/query crops remain at the tail with their supervision aligned. This cross-frame crop-order shuffling is used only during training: at inference, both variants receive the same crops in the same order, encoded in their respective frame-wise or flat representations, without additional shuffling. This comparison therefore tests the combined effect of frame markers, turn boundaries, and training-time context ordering, rather than isolating any one component.

\begin{table}[t]
\centering
\small
\setlength{\tabcolsep}{3pt}
\begin{tabular}{@{}ccccc@{}}
\toprule
CF & TP & Frame-wise & Flat/shuffled & Paired $\Delta$ \\
\midrule
\multicolumn{5}{c}{SportsMOT} \\
0 & F & $82.56 \pm 0.14$ & $82.43 \pm 0.37$ & $0.12 \pm 0.29$ \\
0 & L & $86.10 \pm 0.20$ & $83.87 \pm 0.71$ & $2.24 \pm 0.57$ \\
1 & L & $89.03 \pm 0.10$ & $85.13 \pm 0.58$ & $3.89 \pm 0.51$ \\
2 & L & $89.92 \pm 0.31$ & $86.26 \pm 0.37$ & $3.66 \pm 0.15$ \\
3 & L & $90.67 \pm 0.11$ & $87.56 \pm 0.48$ & $3.12 \pm 0.40$ \\
\midrule
\multicolumn{5}{c}{DanceTrack} \\
0 & F & $36.52 \pm 0.06$ & $35.44 \pm 0.41$ & $1.09 \pm 0.39$ \\
0 & L & $40.99 \pm 0.23$ & $38.46 \pm 0.88$ & $2.54 \pm 0.94$ \\
1 & L & $43.39 \pm 0.08$ & $39.47 \pm 0.87$ & $3.92 \pm 0.92$ \\
2 & L & $44.04 \pm 0.65$ & $39.87 \pm 0.96$ & $4.17 \pm 1.55$ \\
3 & L & $43.57 \pm 0.40$ & $39.91 \pm 0.76$ & $3.67 \pm 0.89$ \\
\midrule
\multicolumn{5}{c}{SoccerNet-GSR} \\
0 & F & $80.04 \pm 0.21$ & $80.57 \pm 0.20$ & $-0.53 \pm 0.32$ \\
0 & L & $83.16 \pm 0.29$ & $81.82 \pm 0.13$ & $1.35 \pm 0.29$ \\
1 & L & $85.93 \pm 0.39$ & $83.29 \pm 0.06$ & $2.63 \pm 0.42$ \\
2 & L & $86.60 \pm 0.34$ & $84.05 \pm 0.14$ & $2.55 \pm 0.40$ \\
3 & L & $86.64 \pm 0.23$ & $84.25 \pm 0.15$ & $2.38 \pm 0.23$ \\
\bottomrule
\end{tabular}
\caption{Input-format ablation: mean $\pm$ sample standard deviation of mAP (\%) across three training seeds. Paired differences are computed as frame-wise minus flat/shuffled within each seed before averaging.}
\label{tab:input-format}
\end{table}

Table~\ref{tab:input-format} shows higher frame-wise mAP on every dataset under all TP=L settings. At CF=1, TP=L, the paired improvements are 3.89 points on SportsMOT, 3.92 on DanceTrack, and 2.63 on SoccerNet-GSR. Under isolated-crop inference, differences are smaller and mixed, including a 0.53-point decrease on SoccerNet-GSR. These results support preserving frame structure when learning to use contextual observations, while leaving the individual contributions of markers, boundaries, and ordering unresolved.

\begin{table}[t]
\centering
\small
\setlength{\tabcolsep}{3pt}
\begin{tabular}{@{}ccccc@{}}
\toprule
CF & TP & SportsMOT & DanceTrack & SN-GSR \\
\midrule
0 & L & $86.34 \pm 0.11$ & $40.60 \pm 0.12$ & $82.68 \pm 0.06$ \\
1 & L & $88.81 \pm 0.22$ & $43.31 \pm 0.20$ & $85.43 \pm 0.14$ \\
\bottomrule
\end{tabular}
\caption{Inference crop-order robustness: mAP (\%), mean $\pm$ sample standard deviation over five shuffle seeds for one fixed training-seed-42 checkpoint. These deviations measure input-order sensitivity, not training-seed variability.}
\label{tab:inference-crop-shuffle}
\end{table}

\subsection{Inference Crop-Order Robustness}
We separately test whether retrieval depends on a particular within-frame crop order. We fix the Qwen3-VL-Embedding-2B checkpoint trained with seed 42 and evaluate five inference shuffle seeds (0--4) under CF=0, TP=L and CF=1, TP=L. Each frame's crops are independently permuted while preserving frame order and keeping the crop being embedded last in its own frame. The same shared context ordering is used for query and gallery encoding within a run. CF=1 uses the first sampled non-query frame; both configurations retain the main protocol's first-three-non-query-frame gallery exclusion. Thus, these runs use the same crops and ranking population, changing only their within-frame input order. Unlike the context-frame-selection experiment, this experiment does not exclude the union of first, middle, and last candidate frames.

Table~\ref{tab:inference-crop-shuffle} shows limited mAP variation across the tested permutations, with standard deviations of 0.06--0.22 percentage points. Adding one cross-frame context frame improves mean mAP by 2.47 points on SportsMOT, 2.71 on DanceTrack, and 2.75 on SN-GSR (computed before rounding). These results indicate that the observed contextual benefit is not restricted to a single within-frame ordering.

\section{Temporal Aggregation References}
ContextReID uses observations from multiple frames, raising the question of whether its gains can be obtained simply by pooling independently encoded crops rather than adapting embeddings to the supplied context. We therefore evaluate post-hoc temporal aggregation with both label-free nearest-crop associations and ground-truth identity associations. The former provides a comparison without inference-time identity labels, while the latter examines the benefit of correctly associated temporal evidence under privileged supervision.

We reuse the trained SOLIDER and Qwen3-VL-Embedding-2B checkpoints without further training; Qwen3-VL-Embedding-2B features are extracted under CF=0, TP=F. Qwen3-VL-Embedding-2B averages seeds 42, 43, and 44, while SOLIDER averages seeds 42 and 44, retaining the main-paper seed exclusion.

\paragraph{Association and pooling.}
For each target crop, label-free nearest-crop pooling independently selects the maximum-cosine crop from each support frame using the original target embedding. It averages the unit-normalized target and selected features with equal weight, then normalizes the result. No identity labels, confidence threshold, or one-to-one assignment are used. Query and gallery crops are processed independently; gallery aggregation never uses the query frame. The oracle reference instead pools only crops with the same ground-truth sequence-local identity. Missing observations are skipped without searching beyond the selected window; if none is available, the original target embedding is retained. Oracle associations provide privileged inference-time information that ContextReID does not require.

\begin{table}[t]
\centering
\footnotesize
\setlength{\tabcolsep}{2pt}
\begin{tabular}{@{}lcccc@{}}
\toprule
Aggregation & $W$ & SportsMOT & DanceTrack & SN-GSR \\
\midrule
\multicolumn{5}{c}{SOLIDER} \\
Single crop & -- & $81.79 \pm 0.15$ & $26.84 \pm 0.76$ & $82.41 \pm 0.23$ \\
\hdashline[1pt/1pt]
\multirow{3}{*}{Nearest crop} & 1 & $74.91 \pm 0.22$ & $23.81 \pm 1.50$ & $78.03 \pm 0.06$ \\
 & 2 & $77.24 \pm 0.46$ & $23.46 \pm 1.90$ & $78.54 \pm 0.35$ \\
 & 3 & $79.00 \pm 0.39$ & $23.86 \pm 1.27$ & $78.54 \pm 0.92$ \\
\hdashline[1pt/1pt]
\multirow{3}{*}{Oracle ID} & 1 & $94.63 \pm 0.14$ & $88.25 \pm 0.00$ & $93.78 \pm 0.03$ \\
 & 2 & $96.17 \pm 0.06$ & $91.88 \pm 0.01$ & $94.34 \pm 0.12$ \\
 & 3 & $97.71 \pm 0.06$ & $91.83 \pm 0.02$ & $94.53 \pm 0.12$ \\
\midrule
\multicolumn{5}{c}{Qwen3-VL-Embedding-2B} \\
Single crop & -- & $82.56 \pm 0.14$ & $36.52 \pm 0.06$ & $80.04 \pm 0.21$ \\
\hdashline[1pt/1pt]
\multirow{3}{*}{Nearest crop} & 1 & $75.47 \pm 0.32$ & $34.50 \pm 0.95$ & $75.57 \pm 0.46$ \\
 & 2 & $77.26 \pm 0.35$ & $35.08 \pm 1.03$ & $75.55 \pm 0.49$ \\
 & 3 & $79.86 \pm 0.61$ & $35.49 \pm 0.57$ & $75.78 \pm 0.51$ \\
\hdashline[1pt/1pt]
\multirow{3}{*}{Oracle ID} & 1 & $95.41 \pm 0.20$ & $88.51 \pm 0.01$ & $94.35 \pm 0.16$ \\
 & 2 & $96.68 \pm 0.20$ & $92.19 \pm 0.05$ & $94.68 \pm 0.16$ \\
 & 3 & $97.59 \pm 0.10$ & $92.14 \pm 0.06$ & $94.82 \pm 0.17$ \\
\bottomrule
\end{tabular}
\caption{Temporal aggregation using the first $W$ sampled non-query frames. mAP scores are mean $\pm$ sample standard deviation (\%) over three Qwen3-VL-Embedding-2B seeds and two SOLIDER seeds. Oracle ID uses ground-truth inference-time associations and is a privileged reference, not a label-free baseline. All rows share the matched gallery for this table.}
\label{tab:temporal-first}
\end{table}

\begin{table}[t]
\centering
\footnotesize
\setlength{\tabcolsep}{2pt}
\begin{tabular}{@{}lcccc@{}}
\toprule
Aggregation & $W$ & SportsMOT & DanceTrack & SN-GSR \\
\midrule
\multicolumn{5}{c}{SOLIDER} \\
Single crop & -- & $80.97 \pm 0.25$ & $26.02 \pm 0.75$ & $81.43 \pm 0.14$ \\
\hdashline[1pt/1pt]
\multirow{3}{*}{Nearest crop} & 1 & $78.78 \pm 0.10$ & $24.74 \pm 0.14$ & $79.29 \pm 0.35$ \\
 & 2 & $78.38 \pm 0.83$ & $23.85 \pm 0.57$ & $78.79 \pm 0.18$ \\
 & 3 & $78.25 \pm 0.67$ & $24.31 \pm 1.42$ & $77.99 \pm 0.25$ \\
\hdashline[1pt/1pt]
\multirow{3}{*}{Oracle ID} & 1 & $89.03 \pm 0.33$ & $30.70 \pm 0.75$ & $87.88 \pm 0.16$ \\
 & 2 & $92.10 \pm 0.25$ & $34.88 \pm 0.28$ & $90.07 \pm 0.08$ \\
 & 3 & $93.26 \pm 0.19$ & $38.74 \pm 0.61$ & $91.39 \pm 0.18$ \\
\midrule
\multicolumn{5}{c}{Qwen3-VL-Embedding-2B} \\
Single crop & -- & $81.83 \pm 0.27$ & $36.12 \pm 0.04$ & $79.06 \pm 0.28$ \\
\hdashline[1pt/1pt]
\multirow{3}{*}{Nearest crop} & 1 & $79.38 \pm 0.54$ & $35.67 \pm 0.71$ & $77.34 \pm 0.40$ \\
 & 2 & $79.12 \pm 0.24$ & $35.64 \pm 0.59$ & $77.13 \pm 0.27$ \\
 & 3 & $79.02 \pm 0.45$ & $36.16 \pm 0.52$ & $76.63 \pm 0.62$ \\
\hdashline[1pt/1pt]
\multirow{3}{*}{Oracle ID} & 1 & $90.64 \pm 0.36$ & $42.60 \pm 0.58$ & $87.63 \pm 0.07$ \\
 & 2 & $93.34 \pm 0.15$ & $46.65 \pm 0.48$ & $90.38 \pm 0.08$ \\
 & 3 & $94.49 \pm 0.28$ & $49.90 \pm 0.47$ & $92.03 \pm 0.05$ \\
\bottomrule
\end{tabular}
\caption{Temporal aggregation using each target's preceding $W$ eligible sampled frames. mAP scores are mean $\pm$ sample standard deviation (\%) over three Qwen3-VL-Embedding-2B seeds and two SOLIDER seeds. Oracle ID uses ground-truth inference-time associations and is a privileged reference, not a label-free baseline. All rows share the matched gallery for this table.}
\label{tab:temporal-previous}
\end{table}

\begin{figure*}[t]
\centering
\includegraphics[width=\textwidth]{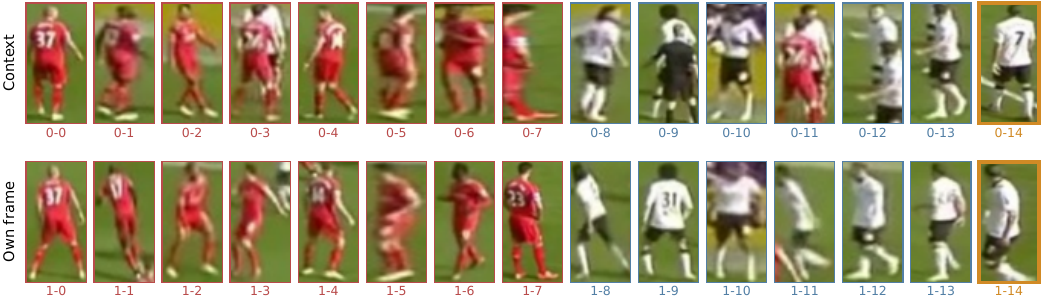}
\caption{All player crops in the attention case study. The top row is the cross-frame context, supplied only under CF=1; the bottom row is the target's own frame. IDs 0--7 and 8--14 form the two teams. Gold borders highlight target identity 14 in both frames, with the own-frame crop $1$-14 serving as the embedding target. Columns are identity-aligned for display only.}
\label{fig:attention-crops}
\end{figure*}

\paragraph{Frame policies and matched evaluation.}
The fixed-first policy uses the first $W\in\{1,2,3\}$ sampled non-query frames as shared support. All settings exclude the first three non-query frames from gallery ranking, matching the main protocol. The preceding-frame policy uses each query or gallery target's own preceding up-to-$W$ eligible sampled frames at approximately 1 FPS. Targets with insufficient history use the available frames. In addition to the main exclusions, the query's preceding three sampled frames are excluded from ranking for every window and for the single-crop reference, but remain eligible as gallery aggregation history. The query frame is always forbidden in gallery history. Consequently, the preceding-frame results must be compared with their own matched reference rather than directly with the main table.

\paragraph{Results.}
Tables~\ref{tab:temporal-first} and~\ref{tab:temporal-previous} report both policies. Nearest-crop pooling reduces mAP relative to its matched single-crop reference on SportsMOT and SoccerNet-GSR for both encoders at every tested window. DanceTrack results also mostly decline, with only a small increase for Qwen3-VL-Embedding-2B using three preceding frames. Oracle identity associations instead substantially improve retrieval, illustrating the benefit of correctly associated temporal observations. These results show that the tested label-free association-and-pooling scheme does not reproduce ContextReID's gains. In particular, oracle pooling addresses a different, association-informed inference setting.

\section{Attention-Mass Case Study}
We examine how ContextReID distributes attention over supplied player crops using Qwen3-VL-Embedding-2B on one SportsMOT frame pair. Figure~\ref{fig:attention-crops} shows all crops: IDs 8--14 belong to one team and IDs 0--7 to the other; player 14 in the own frame is the target. Labels $0$-$i$ and $1$-$i$ denote player $i$ in the context and own frame, respectively. IDs and team membership are used only for this analysis and visualization, not supplied to the model. Identity-aligned columns in the contact sheet do not represent the permuted model input order.

\begin{figure*}[t]
\centering
\includegraphics[width=\textwidth]{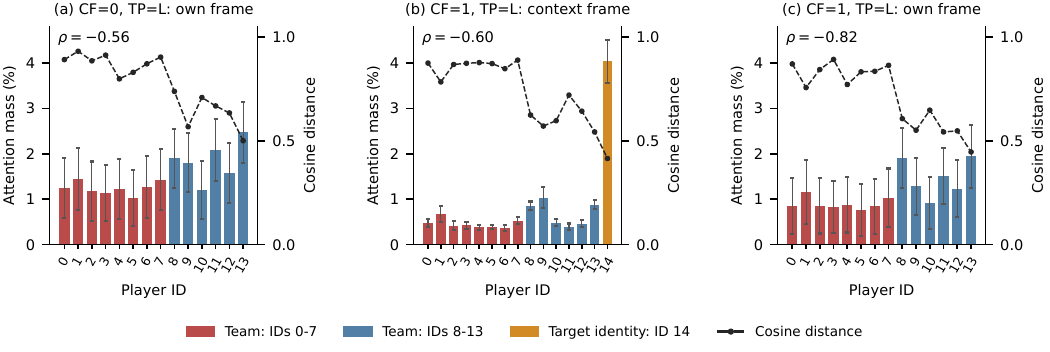}
\caption{Attention-mass distributions under balanced crop orders: (a) CF=0, TP=L over 28 orders; (b,c) CF=1, TP=L over 210 order combinations. Bars show mean attention mass (left axis), with sample standard deviations across orders; dashed curves show mean contextualized-embedding cosine distance from the target (right axis). The target self-pair is excluded. Colors identify the two teams, with gold denoting the context observation of target identity 14. $\rho$ is the Spearman correlation between per-crop mean distance and mean mass in each panel. Order variability is not a confidence interval or training-seed variability.}
\label{fig:attention-mass}
\end{figure*}

\paragraph{Measurement and order controls.}
For each attention head and layer, we take the attention row at the target crop's \texttt{<|vision\_end|>} readout and sum the weights assigned to each crop's 32 visual tokens and its \texttt{<|vision\_end|>} token. We average these masses over heads, layers, and orders without renormalizing over crops; attention to other text and special tokens is not included in the plotted bars. We exclude the own-frame target self-pair. Cosine distances are computed between final crop readouts and the target readout within each order, then averaged by crop identity.

Under CF=0, TP=L, we use 14 cyclic rotations and their 14 reversed counterparts, placing every non-target crop at each available position twice while keeping target 14 last. Under CF=1, TP=L, we independently rotate the 15 context crops and 14 non-target own-frame crops, yielding $15\times14=210$ combinations, again with the target last. This balances positions within each frame group. Error bars are sample standard deviations across these dependent orders, not independent repetitions or confidence intervals.

\paragraph{Appearance-related but non-monotonic attention.}
Figure~\ref{fig:attention-mass}(a) shows a negative relationship between embedding distance and attention mass under CF=0: Spearman $\rho=-0.56$. Same-team players often have smaller distances and attract more attention, but the correspondence is not strict. For example, player 10 receives less attention than player 1 despite having a substantially smaller embedding distance. Under CF=1, the correlations remain negative for both the context frame ($\rho=-0.60$) and other own-frame crops ($\rho=-0.82$). These patterns are consistent with appearance-related attention.

\paragraph{Attention to the cross-frame target observation.}
With CF=1, TP=L, the context observation of player 14 receives $4.03\pm0.48\%$ attention mass, the largest among all non-self crops. The next-largest context-crop mean is $1.04\%$ for player 9. The emphasis on the same identity across frames is consistent with the model recognizing a relevant cross-frame observation without being given its identity or association. However, attention alone does not establish successful identity recognition or a causal contribution to retrieval. This single-target case study reveals some connections between attention and identity. How these patterns generalize across frames, players, and sports remains an open question.

\end{document}